\pdfoutput=1
\documentclass[11pt]{article}

\usepackage[]{ACL2023}

\usepackage{times}
\usepackage{latexsym}

\usepackage[T1]{fontenc}
\usepackage[utf8]{inputenc}

\usepackage{microtype}

\usepackage{inconsolata}
\usepackage{multirow}
\usepackage{booktabs}
\usepackage{amsmath}
\usepackage{ragged2e}
\usepackage{color}
\usepackage{graphicx}
\usepackage{url}
\usepackage{tablefootnote}
\usepackage{subfigure}
\usepackage[bottom]{footmisc}
\usepackage{xcolor}
\definecolor{ypcolor}{rgb}{0.04, 0.73, 0.71}

\title{MultiCoder: Multi-Programming-Lingual Pre-Training for Low-Resource Code Completion}

\author{Zi Gong\textsuperscript{1*}, Yinpeng Guo\textsuperscript{2*}, Pingyi Zhou\textsuperscript{2}, Cuiyun Gao\textsuperscript{1\dag}, Yasheng Wang\textsuperscript{2}, Zenglin Xu\textsuperscript{1\dag} \\
\textsuperscript{1}Harbin Institute of Technology (Shenzhen)\\
\textsuperscript{2}Huawei Noah’s Ark Lab\\
gongzi98521@gmail.com, \{gaocuiyun, xuzenglin\}@hit.edu.cn,\\ \{guo.yinpeng, zhoupingyi, wangyasheng\}@huawei.com
}

\begin{document}
\maketitle
\begingroup\def\thefootnote{*}\footnotetext{Equal contribution}\endgroup
\begingroup\def\thefootnote{\dag}\footnotetext{Corresponding author}\endgroup

\begin{abstract}
    Code completion is a valuable topic in both academia and industry. 
    Recently, large-scale mono-programming-lingual (MonoPL) pre-training models have been proposed to boost the performance of code completion.
    However, the code completion on low-resource programming languages (PL) is difficult for the data-driven paradigm, while there are plenty of developers using low-resource PLs. 
    % Multilingual pre-training is proved to benefit low-resource languages~\cite{mbert, xlm_NEURIPS2019}. 
    On the other hand, there are few studies exploring the effects of multi-programming-lingual (MultiPL) pre-training for the code completion, especially the impact on low-resource programming languages.
    To this end, we propose the \textit{MultiCoder} to enhance the low-resource code completion via MultiPL pre-training and MultiPL Mixture-of-Experts (MoE) layers. We further propose a novel PL-level MoE routing strategy (\textit{PL-MoE}) for improving the code completion on all PLs.
    Experimental results on CodeXGLUE and MultiCC demonstrate that 1) the proposed \textit{MultiCoder} significantly outperforms the MonoPL baselines on low-resource programming languages, and 2) the \textit{PL-MoE} module further boosts the performance on six programming languages.
    In addition, we analyze the effects of the proposed method in details and explore the effectiveness of our method in a variety of scenarios.
\end{abstract}

\section{Introduction}
\label{sec:introduction}
% Code intelligence is a critical research area in software engineering.
% With the rapid growth of deep learning, the field of using deep models to solve code-related tasks has attracted wide attention. 
% Many previous works have proven the effectiveness of deep learning based methods on various tasks associated with programming languages, 
% It covers a wide range of applications, such as code completion~\cite{svyatkovskiy2020intellicode, liu2020multi, izadi2022codefill}, code summarization~\cite{ahmad2020transformer, wu2020code, gong2022source}, code search~\cite{wan2019multi, husain2019codesearchnet} and code clone detection~\cite{tufano2018deep, zhao2018deepsim}. %Previous studies have greatly promoted the advancement of code intelligence. 
% In this paper, we focus on the task of code completion,
Code compleltion is a valuable topic in both academia and industry.
It is an essential and important functionality of current Integrated Development Environments (IDEs)~\cite{svyatkovskiy2020intellicode,izadi2022codefill,amann2016study}. Code completion suggests the subsequent code tokens, such as method calls or object fields, based on the existing code context. This technology facilitates programmers in daily programming activities and can greatly speeds up the software development. 
% For example, Svyatkovskiy \textit{et al.}~\cite{svyatkovskiy2021fast} propose a novel neural reranking model achieves 90\% completion accuracy on the top five suggestions within 8 ms.
% \yun{process [any cite or support?]}.

Recent advances in natural language processing (NLP) and code intelligence fields have revealed the prevalence of  large-scale pre-trained language models, such as BERT~\cite{devlin2019bert}, CodeBERT~\cite{feng2020codebert}, CodeGPT~\cite{lu2021codexglue} and CodeT5~\cite{wang2021codet5}, among which CodeGPT, a decoder-only model, is specifically designed
% In particular, a decoder-only model, named as CodeGPT, has been proposed 
for code completion and other code generation tasks~\cite{lu2021codexglue}.
%In recent years,  make huge progress in , such as . 
%\cite{feng2020codebert} first introduce PLM into code intelligence to address code understanding tasks. 
%Since then, a growing number of code pre-training models have been proposed, they have greatly promoted the development of code intelligence. It is well-know that the pre-training models work well mainly thanks to the availability of large-scale programming languages corpus. 
%In this work, we adopt its decoder-only network structure and autoregressive generation method.We extend the decoder-only generation model to MultiPL field.

In terms of the number of programming languages, there are two pre-training paradigms: mono-programming-lingual (MonoPL) pre-training and multi-programming-lingual (MultiPL) pre-training. 
% The MonoPL models are pre-trained on only one programming language while the MultiPL models are pre-trained on more than one programming languages. 
Although previous MonoPL pre-trained models~\cite{lu2021codexglue, liu2020multi, izadi2022codefill} significantly improve the code completion for specific programming languages, they do not perform well on low-resource programming languages since MonoPL models require language-wise large-scale pre-training datasets. Meanwhile, multilingual pre-training is proved to benefit low-resource languages through cross-lingual knowledge transfer~\cite{zoph2016transfer,conneau2019cross}. 
Hence, we propose to apply MultiPL pre-training in this work.
% By transferring the knowledge from high-resource programming languages, MultiPL models may alleviate the poor performance of code completion in low-resource scenario. 
% the large-scale parameters of a MonoPL model are trained to fit a specific programming language, it requires large-scale pre-training datasets of the programming language. 
% However, due to the diversity of programming languages, many programming languages are of limited data resources. MonoPL pre-training may overfit on the low-resource programming languages, causing poor low-resource code completion performance. %However, the code completion for low-resource programming languages is also important. 
% On the other hand, MultiPL models may improve the code completion on low-resource programming languages with the help of knowledge transfer from high-resource programming languages. Hence, this paper proposes \textit{MultiCoder} to boost low-resource code completion via MultiPL pre-training. 
Nevertheless, the improvement of MultiPL models in low-resource programming languages is still limited. First, as the number of languages and size of dataset increase, the models with fixed capacity are difficult to distinguish the exclusive features of different languages~\cite{arivazhagan2019massively}. 
Second, due to the reduction of capacity for each language, the overall performance tend to be degraded ~\cite{johnson2017google,tan2018multilingual}.
% is limited and may even hurt the performance of high-resource programming languages. Arivazhagan \textit{et al.} observe that as the number of languages with fixed model capacity increases, the positive/negative transfer boundary becomes salient, 
% and the performance of high-resource languages degrade due to a reduction in per-language capacity~\cite{arivazhagan2019massively}. 
% and high-resource languages start to regress due to a reduction in per-task capacity~\cite{arivazhagan2019massively}. 
% Nevertheless, the mixture of multiple programming languages can hurt the performance on high-resource programming languages due to the capacity bottleneck of model parameters~\cite{arivazhagan2019massively}. In other words, the size of the mixed data is too large for a model to well learn. 
% This may cause some exclusive features of different programming languages be discarded.
% \yp{To which extent the parameters are shared across languages determines the magnitude of positive transfer~\cite{baldwin1988transfer} and language interference brought by the capacity bottleneck~\cite{arivazhagan2019massively}}. 
% Thus, effectively expanding the capacity of MultiPL pre-trained models is the key to enhance the performance of low-resource programming languages and achieve more accurate code completion.
% code completion performance and 
% preserve the gains in high-resource programming languages.
% \yun{[@gongzi: modifiy this para.]}

To mitigate the capacity bottleneck of MultiPL pre-trained models, we propose to build models upon the Mixture-of-Experts (MoE) which can scale models up to billions of parameters without a proportional increase in computation~\cite{lepikhin2020gshard,fedus2022switch, gururangan2021demix, kudugunta2021beyond}.
% is widely used to expand the capacity of language models in the NLP field~\cite{lepikhin2020gshard,fedus2022switch, gururangan2021demix, kudugunta2021beyond}.
% Recently, the Mixture-of-Experts (MoE) architecture is widely used to expand the capacity of the language models~\cite{lepikhin2020gshard}.
% MoE architecture consists of a gated network and multiple expert sub-networks, and selectively activates partial parameters in the model for different inputs to participate in the calculation. Therefore, the model can be scaled up to billions of parameters without a proportional increase in computation.
% Previous studies demonstrate the great potential of MoE for MultiPL pre-training~\cite{fedus2022switch, gururangan2021demix, kudugunta2021beyond}. 
%Inspired by these studies, we introduce MoE layers to improve MultiPL code completion. However, a drawback of previous MoE models is that they do not guarantee exclusive experts for each programming language.
%To address the above problems in MultiPL,
%in this work, we adopt a decoder-only network structure and autoregressive generation method. %We extend the decoder-only generation model to MultiPL field.
% Inspired by the successes of MoE for natural language processing~\cite{lepikhin2020gshard,fedus2022switch, gururangan2021demix, kudugunta2021beyond}, 
However, since the parameters in all experts of MoE layers are shared, the existing MoE models~\cite{lepikhin2020gshard,fedus2022switch} fail to guarantee exclusive experts for each programming language, leading to limited performance in MultiPL code completion.
% we introduce MoE layers to improve MultiPL code completion. 
% MoE can selectively activate a part of the parameters in the model for different inputs to participate in the calculation, so that the model can be scaled up to billions of parameters without a proportional increase in computation.
% However, a drawback of previous MoE models is that they do not guarantee exclusive experts for each programming language.
%Recently, the Mixture-of-Experts (MoE) architecture is widely used to expand the capacity of the language models~\cite{lepikhin2020gshard}.
%MoE can selectively activate a part of the parameters in the model for different inputs to participate in the calculation, so that the model can be scaled up to billions of parameters without a proportional increase in computation.
%Previous studies demonstrate the great potential of MoE for MultiPL pre-training~\cite{fedus2022switch, gururangan2021demix, kudugunta2021beyond}. 
%Inspired by these studies, we introduce MoE layers to improve MultiPL code completion. However, a drawback of previous MoE models is that they do not guarantee exclusive experts for each programming language.
%routing strategies do not pay attention to the problem of interference between different languages. In an ideal scenario, we would like to train a single large MultiPL model maximizing the ability of transfer knowledge; meanwhile, we would want to enjoy the benefits of expanding the capacity bottleneck using MoE. 
To explicitly expand the capacity for each programming language, this paper proposes a PL-level MoE routing strategy (\textit{PL-MoE}) that determines exclusive experts for each programming language. Specifically, we assign a group of experts to each programming language according to the data distribution. 
The PL-specific knowledge is learned by the PL-specific parameters in the PL-MoE layer.
%PL-specific parameters are explicitly reserved to improve the capacity for each programming language. 
Meanwhile, the PL-MoE layer also keeps a set of shared experts since not all the tokens are PL-specific. In this way, the PL-MoE is able to learn both the PL-specific and the PL-agnostic knowledge.
%\gz{We also add an additional set of shared experts at each MoE layer to preserve the transferability of common features.}
% In order to consider the knowledge transferability between multiple programming languages, we use dense architecture at the bottom layers of model and add extra shared experts at each MoE layer. 
Furthermore, previous work~\cite{jawahar2019does} shows that pre-trained language models encode a rich hierarchy of linguistic information: surface information at the bottom, semantic information at the top. Thus, we construct our model in the following hierarchical fashion: First, we apply dense architecture at the bottom layers of model to learn transfer and surface information of all programming languages. Second, we apply MoE on the top layers of the model to learn exclusive semantic features of each programming language.
% \yun{[further modify this]}In addition, we apply MoE only on the top layers of the model so that knowledge transfer between programming languages is retained on the bottom layers.

We evaluate the proposed MultiCoder on the CodeXGLUE~\cite{lu2021codexglue} MultiPL code completion tasks and the newly created MultiCC code completion dataset. %The MultiCC dataset consists of six programming languages: Go, Java, JavaScript, Python, PHP and Ruby. 
Furthermore, we conduct extensive experiments to verify and analyze the effectiveness of the MultiCoder in low-resource and cross-domain scenarios. Finally, we discuss the broad application prospects of the MultiCoder in details.

The main contributions of this paper include:
\begin{itemize}
    \item To the best of our knowledge, the \textit{MultiCoder} is the first attempt to improve low-resource code completion via multi-programming-lingual pre-training. %we are the first to explore the use of the MoE architecture in MultiPL pre-training models for the task of code completion.
    \item We propose a novel \textit{PL-MoE} routing strategy which expands the model capacity for each programming language in the MultiPL pre-training.
    \item We create the MultiCC dataset, a MultiPL code completion dataset, which consists of six programming languages. It can promote the research on MultiPL code completion in the software engineering community.
    % It facilitates a more comprehensive study of high-resource and low-resource programming languages.
    \item The proposed MultiCoder is proved effective by the experimental results on CodeXGLUE and MultiCC. A series of further experiments demonstrate the advantage of the model in low-resource and cross-domain scenarios.
\end{itemize}

% The rest of the paper is organized as follows. Section~\ref{sec:background} introduces the background knowledge and motivation of this paper. Section~\ref{sec:methodology} presents the method details of the MultiCoder. Section~\ref{sec:setup} presents the experimental setup and Section~\ref{sec:results} shows the experimental results. Section~\ref{sec:dicussions} discusses the effectiveness and the threats to validity of the proposed approaches. Section~\ref{sec:related} introduces the related work of the research. Finally, Section~\ref{sec:conclusion} summarizes the entire article and discusses the future work.

\begin{figure*}[!ht]
    \centering
    \includegraphics[width=0.95\textwidth]{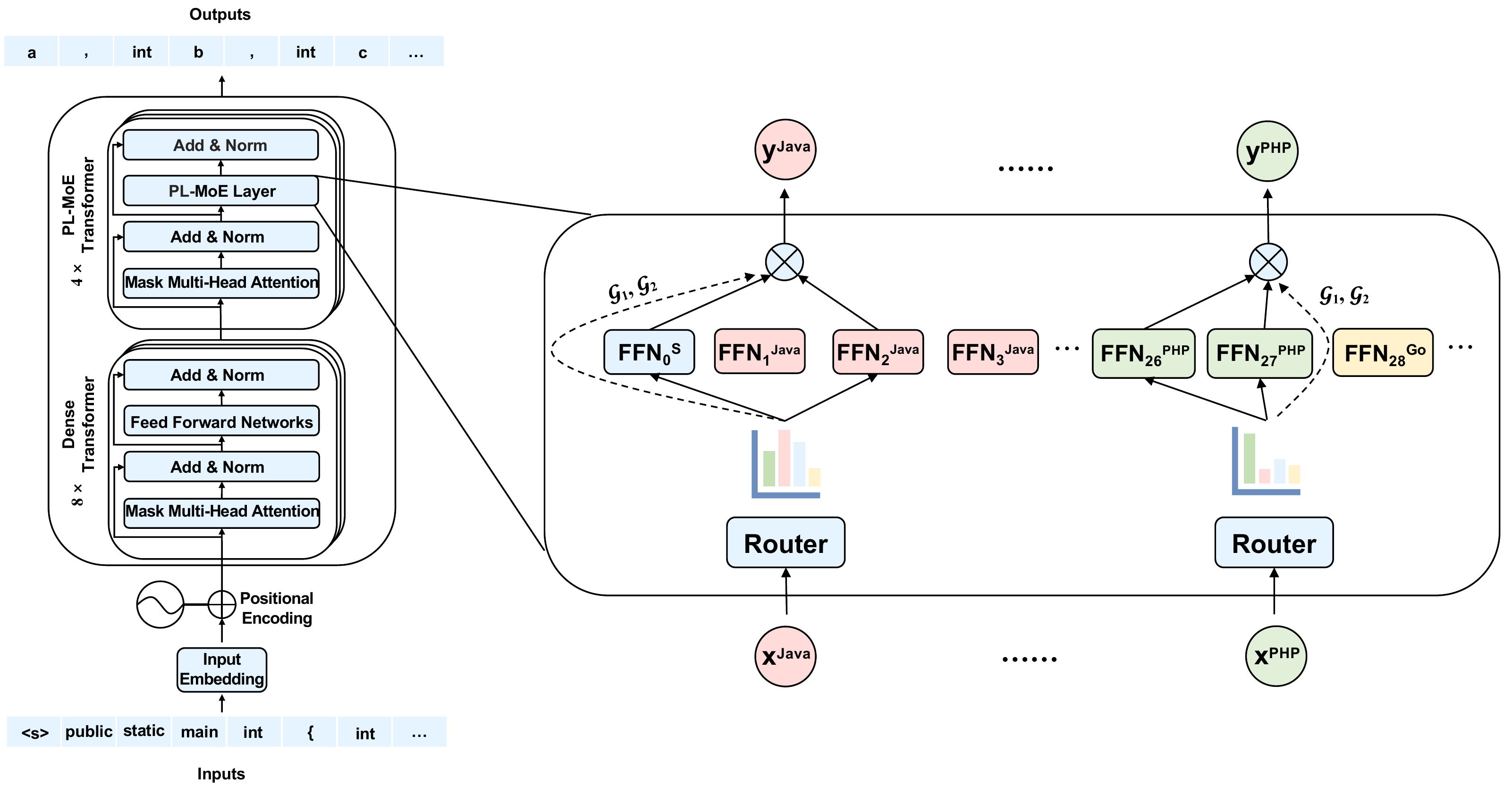}
    \caption{The illustration of the overall architecture of MultiCoder (left part) and PL-MoE layer (right part).
    % and the right part is the illustration of PL-MoE layer.
    }
    \label{fig:PL-MoE}
\end{figure*}
\section{Methodology}
\label{sec:methodology}

% As mentioned in Section~\ref{sec:introduction}, insufficient pre-training data for low-resource programming languages limits the performance of pre-trained models for code completion on these languages.
% As mentioned in Section~\ref{sec:introduction}, insufficient pre-training data for low-resource programming languages limits the code completion performance on these languages.
The previous works~\cite{conneau2019cross, arivazhagan2019massively} have shown the potential of MultiPL pre-training to benefit low-resource programming languages. Inspired by this work, in this paper, we explore the MultiPL pre-training paradigm for low-resource code completion.
Besides, we choose the decoder-only model as our backbone because of its outstanding generation ability~\cite{radford2018improving}.

\subsection{Multi-Programming-Lingual Pre-Training}
% MultiCoder-dense is a MultiPL variant of GPT-2~\cite{radford2019language} that pre-trains the decoder-only language model with multiple programming languages.
The CodeGPT is a variant of GPT-2~\cite{radford2019language} that pre-trains the decoder-only language model with a single programming language. The MultiCoder first expands the CodeGPT via pre-training on multiple programming languages.
%We propose a naive version of MultiCoder called MultiCoder-dense. 
%It uses the conventional dense-activated GPT-2~\cite{radford2019language} architecture with Byte Pair Encoding (BPE) tokenization~\cite{sennrich2016neural}. 
It follows the dense-activated network architecture and causal language modeling of GPT-2, which predicts the next token ${x}_{i}$ condition on previous tokens $\{{x}_{0}, {x}_{1}, \ldots {x}_{i-1}\}$.
The training objective is to minimize the negative logarithm prediction probability:
\begin{equation}
\mathcal{L}=-\sum_{i} \log P\left(x_{i} \mid x_0, x_1, \ldots x_{i-1}\right),%x_{<i}\right)% \rightarrow \max
\label{eqa:4}
\end{equation}
besides, we adopt the Byte Pair Encoding (BPE) tokenization~\cite{sennrich2016neural}.

%In addition to lowering operational costs, 
%In our preliminary experiments as Table~\ref{tab:multicc} shows, although MultiPL pre-training benefits low-resource programming languages compared to MonoPL pre-trianing, it hurts the high-resource programming languages. 
% Although the {MultiPL pre-training} benefits the low-resource programming languages compared to the MonoPL pre-trianing, it hurts the performance on other programming languages due to the capacity bottleneck of model parameters~\cite{johnson2017google,tan2018multilingual}.
%through joint training and consequent positive transfer from high-resource programming languages. 
%Unfortunately, this simultaneously would 
% which are even underperforms the counterpart of MonoPL model, 
% due to interference and capacity bottleneck~\cite{tan2018multilingual}.
% To alleviate this problem, we expand the model capacity of the MultiCoder. % added for example
% Inspired by~\cite{tan2018multilingual}, we expand the model capacity to encourage the MultiCoder robustly outperform the MonoPL baseline on all languages.

\subsection{Mixture-of-Experts (MoE)}
To expand the model capacity, the MultiCoder introduces the \texttt{mixture of experts} (MoE) to its feed-forward layers. The Switch Transformer layers~\cite{fedus2022switch} are adopted as the basic implementation. Specifically, MultiCoder takes the code token ${x}$ as input and then routes it to the best expert, selected from a set $\left\{FFN_{i}(x)\right\}_{i=1}^{N}$ of N experts. The router variable $W_{r}$ yields the logits $h_{x} = W_{r} \cdot {x}$ which are normalized via a softmax distribution over the available N experts at that layer. The gate score for expert $i$ is given as:
\begin{equation}
    g_{i}(x)={\frac{e^{h(x)_{i}}}{\sum_{j}^{N} e^{h(x)_{j}}}}.
\end{equation}
The top-$1$ gate value is selected for routing the token $x$. Then the output computation is a linearly weighted combination of the expert's computation of the token and its gate value, as follows:
\begin{equation}
    y={g_{i}(x) FFN_{i}(x)}.
\end{equation}
% \begin{equation}
%     \mathcal{G}_{i, E}=\operatorname{GATE}\left(x_{i}\right)
%     \label{eqa:1}
% \end{equation}
% \begin{equation}
%     \mathrm{FFN}_{e}\left(x_{i}\right)=w o_{e} \cdot \operatorname{ReLU}\left(w i_{e} \cdot x_{i}\right)
%     \label{eqa:2}
% \end{equation}
% \begin{equation}
%     y_{i}=\sum_{e=1}^{E} \mathcal{G}_{i, e} \cdot \mathrm{FFN}_{e}\left(x_{i}\right)
%     \label{eqa:3}
% \end{equation}
% where $x_{i}$ is the input token at position $i$ to the MoE layer. The vector $\mathcal{G}_{i, E}$ is computed by the gating network. And it has one non-negative value for each expert, which determines the assignment of experts. And the ${FFN}_{e}$ (an expert) refers to a 2-layer fully-connected network using ReLU activation function, $w i_{e}$ and $w o_{e}$ refer to the input and output projection matrices for the $e$-th expert. 
% Following~\cite{fedus2022switch}, we add an load balancing loss to encourage the experts to be trained with same number of tokens. 
More details about the Switch Transformer are given in~\cite{fedus2022switch}.

% Despite MoE expands the model capacity, it does not take into account the exclusive features of different programming languages. In details, \yp{all the programming languages share the identical experts on each MoE layer. 
% The routing strategy of MoE is trained for token distribution rather than for different programming languages.} 
% % This issue is especially significant on high-resource languages. 
% Therefore, the MoE layers should be improved to guarantee the exclusive features of different programming languages. 

\subsection{Programming Language Specific Experts}
In order to protect the exclusive features of different programming languages, the MultiCoder proposes a novel \texttt{programming language specific experts} (PL-MoE). % PL-level MoE routing strategy. 
% PL-exclusive experts
%To benefit from transfer knowledge, we deploy the dense architecture at the bottom layers of model. 
%We deploy sparse PL-MoE architecture at the top layers to break through the capacity bottleneck and alleviate the interference of different programming languages. 
The overall architecture of MultiCoder and the illustration of the routing strategy in PL-MoE layer are shown in the Figure~\ref{fig:PL-MoE}. The tokens of a programming language are routed to the PL-specific experts in the PL-MoE layer. The MultiCoder assigns a group of experts to each PL according to the distribution of PL's resources:
$\{FFN_{i}^{p}(x^{p})\}_{i=1}^{E^{p}}$
%\begin{equation}
%    \{FFN_{e}^{p}(x_{i}^{p})\}_{e=1}^{E^{p}}
%    \label{eqa:5}
%\end{equation}
, where $p$ refers to the $p$ programming language ($p=Java, Python ...$), $x^{p}$ is a token in a sample of $p$, $E^{p}$ is the set of experts assigned to $p$, and $FFN_{i}^{p}$ is an expert assigned to $p$.

% In addition, it is difficult for the model to share knowledge across programming languages with completely isolated PL-MoE parameters.
%if the data from different PLs are completely isolated at the MoE layer, the ability of the model to learn transfer knowledge will be weakened. 
% To improve the trasferrability, we add 
The MultiCoder further proposes a set of \texttt{shared experts} $E^{S}$ to each PL-MoE layer to share language-agnostic features. Thus the gate score of $E^{p}$ and $E^{S}$ is given by:
\begin{equation}
    \mathcal{G}_{E^{p}+E^{S}}=\operatorname{GATE}\left(x^{p}\right).
    \label{eqa:6}
\end{equation}
% Then the token $x_{i}^{p}$ is routed to the top-$k$ PL-exclusive experts $E_{\{0,\ldots,k\}}^p$, ranked by the gate scores. 
Then the token $x^{p}$ is routed to the top-$k$ experts ranked by the gate scores. 
The output of the PL-MoE layer is the gate-weighted summation of the experts' outputs on the token,
\begin{equation}
\begin{aligned}
    \label{eqa:7}
    y^{p}&=\sum_{i \in \phi} \mathcal{G}_{i} \cdot \mathrm{FFN}_{i}\left(x^{p}\right), \\
    % \phi&=E_{\{0,\ldots,k\}}^p \cup E^{shared}
    % \phi&=top_{k}(\mathcal{G}_{i, E^{p}+E^{shared}})
\end{aligned}
\end{equation}
where $\phi$ refers to the set of selected top-$k$ indices from $E^{p}$ and $E^{S}$. In this paper, we take the value of $k$ as 2.
It is worth noting that those experts which do not belong to $\phi$ will not be activated. The purpose of the sparse activation is to reduce the computational overhead. For example, as shown in Figure~\ref{fig:PL-MoE}, when the Java token representation is feed into PL-MoE layer, the router will compute the gate score only for $E^{Java}$ and $E^{S}$. In this case, the $E^{S}$ and $E_{2}^{Java}$ are selected, which are the top-$2$ gate values. Finally, the output is: \\
\begin{equation}
    \mathcal{G}_{1} \cdot FFN^{S}(x^{p}) + \mathcal{G}_{2} \cdot FFN_{2}^{Java}(x^{p}). 
    \nonumber
\end{equation}
% \begin{equation}
% \operatorname{GATE}(x_i^{Java}) \cdot FFN^{shared} + \operatorname{GATE}(x_i^{Java}) \cdot FFN_{2}^{Java} \nonumber
% \end{equation}

\section{Experimental Setup}
\label{sec:setup}
In this section, we design extensive experiments to verify the effectiveness of the MultiCoder. We adopt the GPT-2 as the backbone to conduct the experiments, the model settings are shown in Table~\ref{tab:settings}. We first evaluate the MultiCoder on code completion of CodeXGLUE~\cite{lu2021codexglue} and MultiCC. Then we explore the low-resource performance of the MultiCoder. Finally, we verify the generalization ability of the MultiCoder with cross-domain code completion. %, which is a more challenge scenario. 

\subsection{Models}
% \item \textbf{LSTM} is a 12 layers LSTM model. It directly fine-tunes on the downstream code completion task with BPE tokenization.
% \item \textbf{Transformer} consists of 12 layers Transformer architecture and directly fine-tunes on the downstream code completion task without any pre-training.
% \item \textbf{GPT-2} is a Transformer-based model pre-trained on natural language corpus. 
% \item \textbf{CodeGPT} are Transformer-based MonoPL models, which are pre-trained respectively on Python and Java corpus. The model architecture and training objectives are same with GPT-2, which consists of 12 layers of Transformer decoders. 
% We do not choose the CodeGPT-adapted as our baseline because it is pre-trained on text corpus in advance.
\noindent\textbf{CodeGPT} are a series of 12-layers GPT-2 models that are pre-trained respectively on 6 programming languages from the CodeSearchNet~\cite{husain2019codesearchnet} and Pytorrent~\cite{bahrami2021pytorrent}. For fair comparison, we reproduce the CodeGPT with identical corpora and training hyper-parameters to the MultiCoder, since~\cite{lu2021codexglue} does not provide the pre-training details of the CodeGPT.

\noindent\textbf{MultiCoder} is a 12-layers GPT-2 based model where the top 4 layers incorporates PL-MoE routing strategy and applies a shared expert. It is pre-trained on the mixed data of the 6 programming languages. The following ablated variants are also included in the experiments.
\begin{itemize}
    \item \textbf{Without shared expert} is an ablated variant of the MultiCoder without adopting the \texttt{shared expert}.
    \item \textbf{Without PL-MoE} is an ablated variant of the MultiCoder without adopting the \texttt{programming language specific experts}. % 12-layers GPT-2 based model where the top 4 layers are deployed with Switch transformer~\cite{fedus2022switch}. Its pre-training data is identical to that of the MultiCoder-dense.
    \item \textbf{Without MoE} is an ablated variant of the MultiCoder without adopting the \texttt{mixture of experts}. % of the identical 12-layers dense-activated GPT-2 architecture to the MonoGPTs. However, it is pre-trained on the mixed data of the 6 programming languages.
    % The MultiCoder-MoE-PL has two variants which are MultiCoder-MoE-PL-top1 and MultiCoder-MoE-PL-top2. The difference between them is the number of experts selected at each time of routing.
\end{itemize}

\subsection{Pre-training}
\subsubsection{Datasets}
\label{sec:dataset_pt}
% \begin{table}[htb]\normalsize
%     \centering
%     \setlength{\tabcolsep}{3pt}
%     \renewcommand{\arraystretch}{1.1}
%     \caption{Data statistics about the full pre-training dataset.}
%     \label{tab:pre-training data}
%     \begin{tabular}{lccc}
%     \toprule
%     Language & \#Samples (CSN) & \#Samples (gongzi) & \#Tokens (B, pingyi) \\
%     \midrule
%     Go & 726,768 & 1,767,287 & 0.6 \\
%     Java & 1,569,889 & 3,106,256 & 1.7 \\
%     JavaScript & 1,857,835 & 2,293,073 & 2.2 \\
%     PHP & 977,821 & 2,851,370 & 1.3 \\
%     Python & 1,156,085 & 8,257,401 & 4.3 (1.7+2.6) \\
%     Ruby & 164,048 & 327,999 & 0.1 \\
%     \bottomrule
%     \end{tabular}
% \end{table}
\begin{table}[t]\small
    \centering
    \setlength{\tabcolsep}{6pt}
    \renewcommand{\arraystretch}{1.1}
    \caption{Data statistics of the full pre-training dataset. M: millions, B: billions.}
    \label{tab:pre-training data}
    \begin{tabular}{lccc}
    \toprule
    Language & Samples Num (M) & Tokens Num (B) \\
    \midrule
    Go & 1.77 & 0.62 \\ %1,767,287
    Java & 3.11 & 1.66 \\ %3,106,256
    JavaScript & 2.29 & 2.25 \\ %2,293,073
    PHP & 2.85 & 1.28 \\ %2,851,370
    Python & 8.26 & 4.29 \\ % 8,257,401
    Ruby & 0.33 & 0.13 \\ % 327,999
    \bottomrule
    \end{tabular}
\end{table}

\noindent\textbf{Full pre-training dataset.} The data statistics of full pre-training dataset is shown in Table~\ref{tab:pre-training data}. We construct the full pre-training dataset by collecting the corpora including six programming languages from the CodeSearchNet and the Pytorrent~\cite{bahrami2021pytorrent}. Samples for each PL are splitted into train, validation and test sets in a 96\%, 2\% and 2\% respectively. Their train sets are mixed for pre-training. For each \textit{(natural language, programming language)} pair, two samples with reverse direction are constructed. Language identifiers are adopted for each PL and NL, e.g. <python>, <java> and <en> for Python, Java and English Docstring respectively.
    
\noindent\textbf{Low-resource pre-training datasets.} They are Java and Python subsets derived from the full pre-training dataset, aiming to evaluate the models in low-resource scenarios.
% We construct two new pre-training datasets, named Java low-resource pre-training dataset and Python low-resource pre-training dataset respectively.
Take the low-resource Java dataset for example, it reduces the number of the Java samples to be the same to the Ruby's, which is the low-resource PL in the full dataset. The same goes for the Python low-resource dataset.
    
\noindent\textbf{Cross-domain pre-training datasets.} They are two subsets of the full pre-training dataset. We remove the Ruby or the Go samples. Pre-trained models are then respectively fine-tuned on Ruby or Go code completion tasks. In this way, the pre-training domain and the fine-tuning domain are different, producing the cross-domain scenario.

\subsubsection{Implementation Details}
All the models are pre-trained on 8 Nvidia Tesla V100 32G GPUs using the Adam~\cite{kingma2015adam} optimizer. The global batch size is 64. The models are pre-trained for 100,000 steps with 1,000 steps warming up. The learning rate starts from 1.5e-4 with cosine decay. The training uses the mixed precision of FP16 and FP32.
The model architecture details are shown in Appendix~\ref{sec:app_model_configs}. 

% \subsection{Fine-tuning on the CodeXGLUE}
\subsection{Fine-tuning}

\begin{table}[t]\small
    \centering
    \setlength{\tabcolsep}{10pt}
    \renewcommand{\arraystretch}{1.1}
    \caption{Statistics of the CodeXGLUE code completion datasets, in number of samples, K: thousands.}
    \label{tab:codexglue-data}
    \begin{tabular}{lccc}
    \toprule
    Language & Train & Dev & Test \\
    \midrule
    % PY150~\cite{raychev2016probabilistic} & 100K & 5K & 50K \\ 
    % Github Java Corpus~\cite{allamanis2013mining} & 13K & 7K & 8K \\ 
    PY150 & 100K & 5K & 50K \\ 
    Github Java Corpus & 13K & 7K & 8K \\ 
    \bottomrule
    \end{tabular}
\end{table}

\subsubsection{Datasets}
% \ypc
\noindent\textbf{PY150}~\cite{raychev2016probabilistic} and \textbf{Github Java Corpus}~\cite{allamanis2013mining} are code completion tasks from the CodeXGLUE~\cite{lu2021codexglue}, which are designed for Python and Java respectively.

% The MultiCC dataset is built for a more comprehensive study of code completion on high-resource and low-resource programming languages \ypc. 
\noindent\textbf{MultiCC} is the \textit{Multi-programming-lingual Code Completion Dataset} created in this paper, from the full pre-training dataset. It's designed for more programming languages not limited to Python and Java.
The examples of MultiCC are randomly sampled from the validation and test set of the original pre-training dataset, so that they have never been seen by the models before fine-tuning.
% The MultiCC is the combination of the valid and test set of the full pre-training dataset. 
The examples are constructed according to the rules described in Appendix~\ref{sec:app_multicc_rules}.
% We name this dataset multiple programming languages code completion dataset (MultiCC) whose data statistics as shown in Table~\ref{tab:multicc-data}. 
Table~\ref{tab:multicc-data} provides the statistics of the MultiCC dataset.
% We unify the sample number of training sets and test sets for six PL in order to eliminate the imbalance of language resources during the fine tuning phase.
In order to make fair comparisons between programming languages, the MultiCC controls the data size for each PL to be identical.

\subsubsection{Implementation Details}
All the models are pre-trained on the full pre-training dataset before the fine-tuning.
For the PY150, Github Java Corpus and MultiCC tasks, the models are fine-tuned on 8, 2 and 2 GPUs for 50K, 2K and 2K steps respectively.
The global batch sizes are 64, 32 and 4 respectively.
The learning rates are 7e-4, 7e-5 and 5e-5 respectively and linearly decayed.

\begin{table}[t]\small
    \centering
    \setlength{\tabcolsep}{6pt}
    \renewcommand{\arraystretch}{1.1}
    \caption{Statistics of the MultiCC. K: thousands. M: millions.}
    \label{tab:multicc-data}
    \begin{tabular}{lcccc}
    \toprule
    Language & \# Tokens (M)& Train & Dev & Test  \\
    \midrule
    Go & 0.93& \multirow{6}{*}{7K} & \multirow{6}{*}{1K} & \multirow{6}{*}{1K}  \\%932,072 \\
    Java & 1.12& &  &   \\ %1,120,122 \\
    JavaScript & 1.17& &  &   \\ %1,170,674 \\
    PHP & 1.12& &  &   \\ %1,123,394 \\
    Python & 1.23& &  &   \\ %1,225,228 \\
    Ruby & 0.73& &  &   \\ %732,490 \\
    \bottomrule
    \end{tabular}
\end{table}

\begin{table*}[t]\small
    \centering
    \setlength{\tabcolsep}{13pt}
    \renewcommand{\arraystretch}{1.1}
    \caption{Comparison of token-level code completion on CodeXGLUE. Results with ``*" are reported by~\cite{lu2021codexglue}. 
    Paired $t$-test between MultiCoder and other models we reproduced are performed, all the $p$-values < 0.0001.}
    \label{tab:codexglue}
    %\resizebox{\textwidth}{!}{
    \begin{tabular}{lcccc|cc}
    \toprule
    \multirow{2}{*}{\pmb{Model}} & \multicolumn{2}{c}{PY150} & \multicolumn{2}{c}{Github Java Corpus} &
    \multicolumn{2}{c}{Overall}\\
    \cline{2 - 7} 
    & Acc & ES & Acc & ES & Acc & ES \\
    \midrule
    % \multirow{5}{*}{Baselines} &
    LSTM~\cite{lu2021codexglue} & 61.94\scriptsize{*} & - & 58.92\scriptsize{*} & - & 60.43\scriptsize{*} & - \\
    Transformer~\cite{lu2021codexglue} & 74.48\scriptsize{*} & - & 65.18\scriptsize{*} & - & 69.83\scriptsize{*} & - \\
    GPT-2~\cite{lu2021codexglue} & 75.90\scriptsize{*} & - & 75.40\scriptsize{*} & - & 75.65\scriptsize{*} & - \\ % 83.37 , 79.73  
    CodeGPT~\cite{lu2021codexglue} & 76.58\scriptsize{*} & - & 76.79\scriptsize{*} & - & 76.69\scriptsize{*} & - \\ % 83.78 , 80.43 
    CodeGPT (ours) & 76.30 & 83.65 & 75.97 & 79.55 & 76.14 & 81.60 \\
    \midrule
    % \multirow{3}{*}{Ours} & 
    MultiCoder & \pmb{76.66} & \pmb{83.84} & \pmb{77.13} & \pmb{80.64} & \pmb{76.90} & \pmb{82.24} \\
    \hspace{3mm}- Shared expert & 76.49 & 83.71 & 76.81 & 80.43 & 76.65 & 82.07 \\
    \hspace{3mm}- PL-MoE & 76.46 & 83.69 & 76.84 & 80.42 & 76.65 & 82.06 \\
    \hspace{3mm}- MoE & 76.37 & 83.65 & 76.61 & 80.32 & 76.49 & 81.99 \\
    % \multirow{4}{*}{Ours} & MultiCoder-dense & 76.37 & 83.65 & 76.61 & 80.32  \\
    % & MultiCoder-MoE & 76.46 & 83.69 & 76.84 & 80.42 \\
    % % MultiCoder-MoE-PL-top1 & 76.62 & 83.80 & 76.92 & 80.46 \\
    % % \multicolumn{1}{c}{w/o shared expert} & 76.40 & 83.67 & 76.78 & 80.39 \\
    % Ours & MultiCoder-MoE-PL & \pmb{76.66} & \pmb{83.84} & \pmb{77.13} & \pmb{80.64} \\
    % & \multicolumn{1}{c}{w/o shared expert} & 76.49 & 83.71 & 76.81 & 80.43 \\
    \bottomrule
    \end{tabular}
    %}
\end{table*}

\subsection{Low-Resource Code Completion}
The models' low-resource variants (including CodeGPT and MultiCoder) are pre-trained on the low-resource Java and Python dataset as described in the Section~\ref{sec:dataset_pt}. They are fine-tuned and evaluated on the MultiCC Java and Python corpora. It is worth noting that the number of experts allocated to Java or Python are the same with Ruby.

\subsection{Cross-Domain Code Completion}
The models are pre-trained on the cross-domain pre-training datasets as described in the Section~\ref{sec:dataset_pt}. 
They are fine-tuned and evaluated on the MultiCC Ruby and Go corpora.
% We pre-train the models (including MultiCoder-dense, MultiCoder-MoE-PL-top1 and MultiCoder-MoE-PL-top2) on the four PL pre-training dataset, then we fine-tune and test them on Ruby and Go corpus of MultiCC. 

\subsection{Evaluation Metrics}
The task of token-level code completion is to predict the next token given the context of the previous tokens.
\begin{itemize}
    \item \textbf{Accuracy} Calculate the token-level accuracy of the next token prediction.
    \item \textbf{Edit Similarity\footnote{In the application of code completion, although a generated code snippet is not exactly correct, programmers may accept and edit it. Therefore, the Levenshtein edit similarity is a critical evaluation metric for code completion.}} The Levenshtein distance between two words is the minimum number of single-character edits required to change one word to the other, including insertions, deletions, or substitutions.
\end{itemize}

\section{Experimental Results}
\label{sec:results}

\begin{table*}[!ht]\small
    \centering
    \setlength{\tabcolsep}{3.5pt}
    \renewcommand{\arraystretch}{1.3}
    \caption{Comparison results on MultiCC. Paired $t$-test between MultiCoder and others are performed, all the $p$-values < 0.0001.}
    \label{tab:multicc}
    \begin{tabular}{lcccccccccccc|ccc}
    \toprule
    \multirow{2}{*}{\pmb{Model}} & \multicolumn{2}{c}{Go} & \multicolumn{2}{c}{Java} & \multicolumn{2}{c}{JavaScript} & \multicolumn{2}{c}{PHP} & \multicolumn{2}{c}{Python} & \multicolumn{2}{c}{Ruby} & \multicolumn{2}{c}{Overall} \\
    \cline{2 - 15} 
    & Acc & ES & Acc & ES & Acc & ES  & Acc & ES  & Acc & ES  & Acc & ES  & Acc & ES \\
    \midrule
    CodeGPT (ours) & 74.25 & 82.54 & 70.41 & 77.16 & 66.48 & 74.48 & 73.19 & 79.28 & 64.60 & 77.34 & 55.76 & 71.58 & 67.45 & 77.06 \\
    \midrule
    MultiCoder & \pmb{74.77} & \pmb{82.91} & \pmb{71.47} & \pmb{77.85} & \pmb{66.99} & \pmb{74.85} & \pmb{74.20} & \pmb{79.89} & \pmb{65.29} & \pmb{77.75} & \pmb{57.59} & \pmb{72.39} & \pmb{68.39} & \pmb{77.61} \\
    % \midrule
    \hspace{3mm}- Shared expert & 74.07 & 82.45 & 70.80 & 77.29 & 66.36 & 74.43 & 73.51 & 79.33 & 64.68 & 77.41 & 57.26 & 72.04 & 67.78 & 77.16 \\
    \hspace{3mm}- PL-MoE & 73.80 & 82.16 & 70.57 & 77.00 & 65.93 & 74.20 & 73.34 & 79.00 & 64.49 & 77.22 & 56.78 & 71.49 & 67.49 & 76.85 \\
    \hspace{3mm}- MoE & 73.27 & 81.76 & 70.29 & 76.74 & 65.73 & 73.93 & 72.97 & 78.86 & 64.47 & 77.20 & 56.54 & 71.70 & 67.19 & 76.67 \\
    % \hspace{3mm}- Multilingual & 74.25 & 82.54 & 70.41 & 77.16 & 66.48 & 74.48 & 73.19 & 79.28 & 64.60 & 77.34 & 55.76 & 71.58 & 67.45 & 77.06 \\
    % MultiCoder-MoE-PL-top1 & \pmb{74.84} & \pmb{83.09} & 71.35 & 77.71 & 66.91 & \pmb{74.90} & 73.71 & 79.53 & 65.12 & 77.64 & \pmb{57.76} & \pmb{72.42} & 68.28 & 77.55 \\
    % \multicolumn{1}{c}{w/o shared expert} & 74.05 & 82.53 & 70.87 & 77.49 & 66.20 & 74.40 & 73.41 & 79.32 & 64.48 & 77.33 & 57.30 & 72.23 & 67.72 & 77.22 \\
    \bottomrule
    \end{tabular}
\end{table*}

\subsection{Code completion}

\subsubsection{Results on CodeXGLUE}
As shown in Table~\ref{tab:codexglue}, the MultiCoder significantly outperforms all the baseline models on both languages and both metrics.

It's worth noting that the MultiCoder outperforms the CodeGPT more significantly on the Java corpus (\textit{Accuracy +1.16, Edit similarity +1.09}) than the Python (\textit{Accuracy +0.36, Edit similarity +0.19}). 
The MultiCoder's advantage may come from two reasons.
First, the MultiCoder benefits the low-resource PLs more than the higher one.
The pre-training data scale of Java is around $1/3$ of Python as Table~\ref{tab:pre-training data} shows. Here Java is a relatively low-resource PL compared with Python.
Second, it implies that the MultiCoder has a better generalization ability than the baseline, since its advantage is more significant on smaller fine-tuning dataset. 
More specifically, the Java fine-tuning set (13K) is less than $1/7$ of the Python set, according to Table~\ref{tab:codexglue-data}.
The better generalization ability of MultiCoder is also demonstrated by the cross-domain code completion in Section~\ref{sec:cross-domain}.

Ablation results are also provided in Table~\ref{tab:codexglue}. 
All of the {shared expert}, {PL-MoE} and {MoE} modules are effective.
However, the \textit{shared expert} brings the most improvement.
It reveals that PL-agnostic features are also important in the PL-MoE module, although the PL-MoE is originally designed for PL-specific representation.
We conjecture that there may be a trade-off between PL-agnostic and PL-specific features on the MoE layers, which we left for the future works.

In addition, the MultiCoder surpasses the baselines even entirely removing the MoE modules, which degenerate it to a multi-programming-lingual dense model.
It implies the advantage of MultiPL pre-training over the MonoPL pre-training.

\subsubsection{Results on MultiCC}
As shown in Table~\ref{tab:multicc}, the advantage of the MultiCoder is verified not only on Java and Python but also on Go, JavaScript, PHP and Ruby. Similar to CodeXGLUE, the MultiCoder surpasses again the baseline on all metrics.

The MultiCoder outperforms the CodeGPT on the low-resource language Ruby (\textit{Accuracy +1.83, Edit Similarity +0.81}) much more than on the high-resource language Python (\textit{Accuracy +0.69, Edit Similarity +0.41}).
The law that the MultiCoder is more beneficial to the lower-resource languages also holds true on Java, JavaScript and PHP.
An exception is on Go - another low-resource PL.
We conjecture the reason is that the syntax and literals of Go are similar to Java's, which augments the training of Go.
Cross-PL transfer results in Table~\ref{tab:cross-domain} evidences this conjecture.
The scores of the CodeGPT-Java on Go is \textit{much higher (>10)} than on Ruby, implying that Java is similar to Go.

The ablation results are similar to those on CodeXGLUE, which verifies the effectiveness of the proposed MoE modules again.

\subsection{Low-Resource Code Completion}
% \begin{table*}[htb]\normalsize
%     \centering
%     \setlength{\tabcolsep}{8pt}
%     \renewcommand{\arraystretch}{1.1}
%     \caption{Low-resource in Java and Python.}
%     \label{tab:low-resource}
%     \begin{tabular}{lcccccc}
%     \toprule
%     \multirow{2}{*}{\pmb{Model}} & \multicolumn{2}{c}{Java} & \multicolumn{2}{c}{Python} & \multicolumn{2}{c}{Overall} \\
%     \cline{2 - 7} 
%     & Accuracy & Edit Sim & Accuracy & Edit Sim & Accuracy & Edit Sim \\
%     \midrule
%     MonoGPT & 68.04 & 74.59 & 61.47 & 75.51 & 64.76 & 75.05 \\
%     MultiCoder-dense & 68.81 & 75.19 & 62.38 & 75.86 & 65.60 & 75.53 \\
%     % MultiCoder-MoE-PL-top1 & \pmb{70.09} & \pmb{76.26} & \pmb{63.37} & 76.50 & \pmb{66.73} & \pmb{76.38} \\
%     MultiCoder-MoE-PL & \pmb{70.01} & \pmb{76.18} & \pmb{63.29} & \pmb{76.52} & \pmb{66.65} & \pmb{76.35} \\
%     \bottomrule
%     \end{tabular}
% \end{table*}
\begin{table*}[htb]\small
    \centering
    \setlength{\tabcolsep}{10pt}
    \renewcommand{\arraystretch}{1.1}
    \caption{Low-resource in Java and Python. 
    Paired $t$-test between MultiCoder and others are performed, all the $p$-values < 0.0001.}
    \label{tab:low-resource}
    \begin{tabular}{lcccc|cc}
    \toprule
    \multirow{2}{*}{\pmb{Model}} & \multicolumn{2}{c}{Java} & \multicolumn{2}{c}{Python} & \multicolumn{2}{c}{Overall} \\
    \cline{2 - 7} 
    & Accuracy & Edit Sim & Accuracy & Edit Sim & Accuracy & Edit Sim \\
    \midrule
    CodeGPT (ours) & 68.04 & 74.59 & 61.47 & 75.51 & 64.76 & 75.05 \\
    \midrule
    MultiCoder & \pmb{70.01} & \pmb{76.18} & \pmb{63.29} & \pmb{76.52} & \pmb{66.65} & \pmb{76.35} \\
    % \hspace{3mm}- Shared expert & - & - & - & - & - & - \\
    % \hspace{3mm}- PL-MoE & - & - & - & - & - & - \\
    \hspace{3mm}- Shared expert, PL-MoE, MoE & 68.81 & 75.19 & 62.38 & 75.86 & 65.60 & 75.53 \\
    % MultiCoder-MoE-PL-top1 & \pmb{70.09} & \pmb{76.26} & \pmb{63.37} & 76.50 & \pmb{66.73} & \pmb{76.38} \\
    \bottomrule
    \end{tabular}
\end{table*}
The low-resource experiments on Java and Python aim to further study the effectiveness of MultiCoder on low-resource PLs. 
% The results are shown in Table~\ref{tab:low-resource}, the MultiCoder outperforms CodeGPT on all languages and metrics, with the overall \textit{Accuracy +1.97 and Edit Similarity +1.33}. 
%Besides, the ablation shows that even without MoE modules, the MultiPL pre-training performs better than the MonoPL baseline. The results demonstrate that: (1) the proposed MultiCoder is effective on low-resource PLs; (2) the MultiPL pre-training benefits the low-resource PLs.
Comparing the Table~\ref{tab:low-resource} with Table~\ref{tab:multicc}, the MultiCoder shows stronger effectiveness on the low-resource settings than the original MultiCC.
Take the accuracy on Java for example, the gap between the MultiCoder and the CodeGPT increases \textit{85.8\% (+1.06 -> +1.97)} on the low-resource experiment. The same goes for the other scores.
The improvement clearly reveals the effectiveness of the MultiCoder for low-resource PLs. 

The ablation in Table~\ref{tab:low-resource} expands the above conclusion to the MultiPL pre-training that is without MoE. 
Specifically, the ablated MultiCoder variant achieves a \textit{qualitative change} on the low-resource settings from the original MultiCC (Table~\ref{tab:multicc}). The overall difference between the MultiPL pre-training (MultiCoder w/o MoE) and the MonoPL training (CodeGPT) changes from \textit{-0.26 to 0.84 in Accuracy} and from \textit{-0.39 to 0.48 in Edit Similarity}.

In short, the MultiCoder and MultiPL pre-training is very effective for low-resource PLs.

\subsection{Cross-Domain Code Completion}
\label{sec:cross-domain}

\begin{table*}[htb]\small
    \centering
    \setlength{\tabcolsep}{10pt}
    \renewcommand{\arraystretch}{1.1}
    \caption{Cross-Domain Code Completion. 
    Paired $t$-test between MultiCoder and others are performed, all the $p$-values < 0.0001.}
    \label{tab:cross-domain}
    \begin{tabular}{lcccc|cc}
    \toprule
    \multirow{2}{*}{\pmb{Model}} & \multicolumn{2}{c}{Go} & \multicolumn{2}{c}{Ruby} & \multicolumn{2}{c}{Overall} \\
    \cline{2 - 7} 
    & Accuracy & Edit Sim & Accuracy & Edit Sim & Accuracy & Edit Sim \\
    \midrule
    CodeGPT-Java (ours) & 68.57 & 78.54 & 52.17 & 68.22 & 60.37 & 73.38 \\
    % CodeGPT-Python (ours) & 69.03 & 78.68 & 52.81 & 68.92 & 60.92 & 73.80 \\
    \midrule
    MultiCoder & \pmb{70.64} & \pmb{79.86} & \pmb{54.60} & \pmb{69.93} & \pmb{62.62} & \pmb{74.90} \\
    % \hspace{3mm}- Shared expert & - & - & - & - & - & - \\
    % \hspace{3mm}- PL-MoE & - & - & - & - & - & - \\
    \hspace{3mm}- Shared expert, PL-MoE, MoE  & 69.96 & 79.47 & 53.70 & 69.38 & 61.83 & 74.43 \\
    % MultiCoder-MoE-PL-top1 & 70.28 & 79.59 & 53.97 & 69.66 & 62.13 & 74.63 \\
    \bottomrule
    \end{tabular}
\end{table*}
% \begin{table*}[htb]\normalsize
%     \centering
%     \setlength{\tabcolsep}{8pt}
%     \renewcommand{\arraystretch}{1.1}
%     \begin{tabular}{lcccccc}
%     \toprule
%     \multirow{2}{*}{\pmb{Model}} & \multicolumn{2}{c}{Ruby} & \multicolumn{2}{c}{Go} & \multicolumn{2}{c}{Overall} \\
%     \cline{2 - 7} 
%     & Accuracy & Edit Sim & Accuracy & Edit Sim & Accuracy & Edit Sim \\
%     \midrule
%     MultiCoder-dense & 53.70 & 69.38 & 69.96 & 79.47 & 61.83 & 74.43 \\
%     MultiCoder-PL-top1 & 53.97 & 69.66 & 70.28 & 79.59 & 62.13 & 74.63 \\
%     MultiCoder-PL-top2 & \pmb{54.60} & \pmb{69.93} & \pmb{70.64} & \pmb{79.86} & \pmb{62.62} & \pmb{74.90} \\
%     \bottomrule
%     \end{tabular}
%     \caption{Cross-Domain Code Completion.}
%     \label{tab:cross-domain}
% \end{table*}
% We conduct a set of experiments to evaluate MultiCoder in cross-domain scenario, which is a more extreme and challenging low-resource scenario. 
% The results are shown in Table~\ref{tab:cross-domain}. 

Cross-domain code completion is a task to evaluate the generalization ability of the MultiCoder.
According to Table~\ref{tab:cross-domain}, the MultiCoder significantly surpasses the CodeGPT. 
It's due to two strategies: the MultiPL pre-training and the PL-MoE modules. 
The ablated MultiCoder in Table~\ref{tab:cross-domain} is based on the naive MultiPL pre-training. Its better performance over the CodeGPT demonstrates the good generalization ability of MultiPL pre-training.
Additionally, PL-MoE further enhances MultiCoder's generalization ability.

We guess the mixture of multiple PLs enables the knowledge transfer between PLs and augments the training between the PLs.
On the other hand, the MoE module stores the PL-specific knowledge, meanwhile it forces the model's other part to learn more PL-agnostic and general knowledge. 
With the enhanced PL-agnostic knowledge, the MultiCoder generalizes better on new-arriving PLs.
% The MultiCoder outperforms either the CodeGPT baselines or the ablated variant.
% The results demonstrates that the PL generalization ability of MultiCoder-MoE-PL is stronger than MultiCoder-dense. And it further proves that MultiCoder-MoE-PL can learn more transfer knowledge from MultiPL pre-training, in order to benefit for low-resource or even cross-domain code completion.

\subsection{Impact of Different MoE Architectures}
% In this section, we conduct a series of ablation experiments on the settings of PL-MoE, including top-$k$ and shared experts.

% Firstly we focus on top-$k$, which refers to the number of experts selected at each time of routing. We set $k$ to 1 and 2 for experiments, respectively. As shown in Table~\ref{tab:codexglue}-\ref{tab:low-resource}, MultiCoder-MoE-PL-top2 outperforms MultiCoder-MoE-PL-top1 in most high-resource PL. On the other hand, MultiCoder-MoE-PL-top1 performs better in low-resource PL, especially in Table~\ref{tab:low-resource}. In cross-domain code completion, MultiCoder-MoE-PL-top2 significantly outperforms MultiCoder-MoE-PL-top1 on overall. We suspect that the more experts are selected for each routing, the better the performance of cross-domain code completion.

% Then we focus on shared experts, which refers to the number of experts are shared for all PL in each PL-MoE layer. We respectively set it to 0 and 1 for experiments, i.e., with or without a shared expert. 
% The results demonstrate that the MultiCoder is able to learn high-level generalized knowledge at the MoE layer at the top of the model, which can obviously improve the performance of code completion. Therefore, it is not optimal for the MultiCoder-MoE-PL to completely isolate different PL at the PL-MoE layer. 
As shown in Table~\ref{tab:codexglue}-\ref{tab:multicc}, the MultiCoder with the shared expert significantly outperforms its counterpart without the shared expert.
An interesting, the improvement on the low-resource Ruby lags behind the similarly low-resource Go. 
The reason may be that Go has similar PLs, e.g. Java, to transfer knowledge through the shared expert module, while Ruby has not.
On the other hand, it implies the shared expert effectively rebalances the trade-off between transfer or not to transfer knowledge on the PL-MoE layers.

In addition, the PL-MoE module performs better than the naive MoE, even without the shared expert. 
It shows that the PL-exclusive experts is critical.

% The main difference between them is whether different PL are routed separately at the MoE layers. As mentioned above, one of the motivations for MultiCoder-MoE-PL is to mitigate the interference between different PL as much as possible. The experimental results evaluate that the PL-MoE can indeed effectively alleviate this problem.
\section{Discussions}
\label{sec:dicussions}

\subsection{A Closer Look at the Routing Decisions}
\begin{figure}[htb]
    \centering
    \subfigure[PL-MoE]{
    \includegraphics[width=6cm]{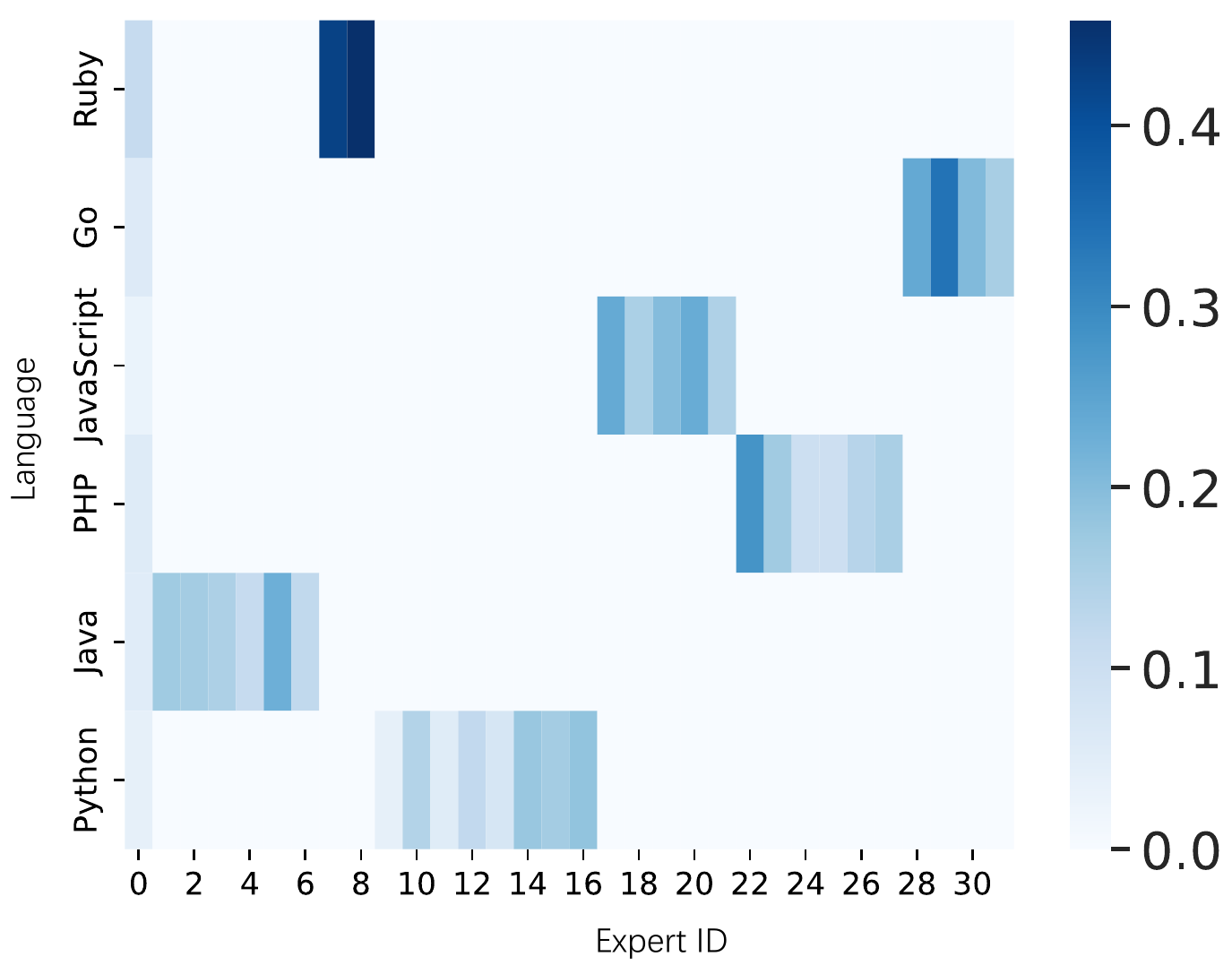}
    \label{routing-pl-moe}
    }
    \\
    % \hspace{-0.8cm}
    \subfigure[MoE]{
    \includegraphics[width=6cm]{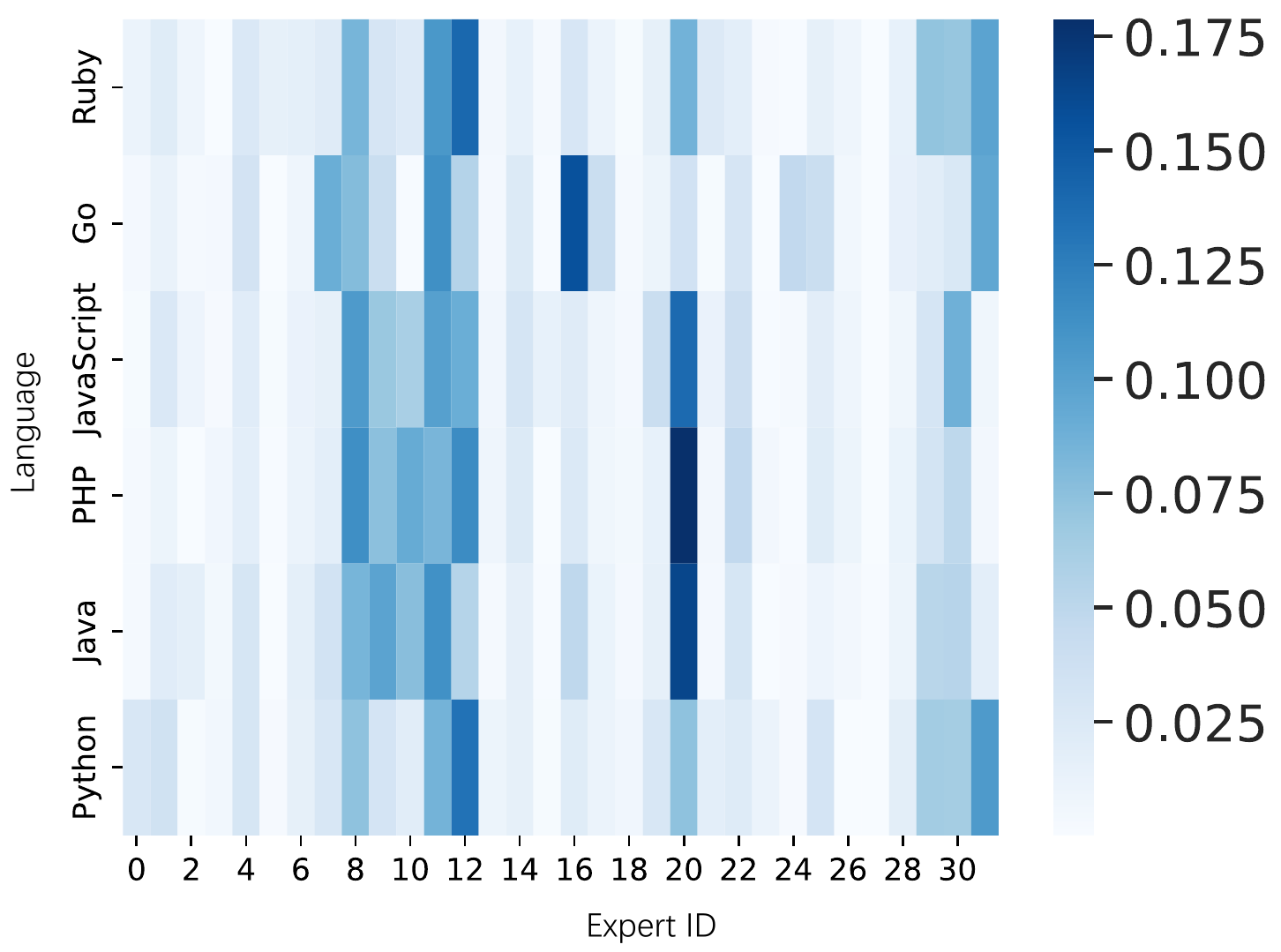}
    \label{routing-moe}
    }
    \caption{Routing decisions of experts (the sum of each row is 1).}% The color annotates the ground truth expert group for each PL. The Expert-0 is the shared expert.}
    \label{fig:routing}
\end{figure}

% Now we analyze the routing decisions made in PL-MoE layers and MoE layers to further motivate our exploration. 
% We use MultiCoder and the variant without PL-MoE to conduct inference on the testing set of MultiCC and log the routing decisions for all the token of each PLs. 
% Then we 
We test on MultiCC and plot the expert distributions on the six PLs as shown in Figure~\ref{fig:routing}.
% For clarity, the expert distribution for each PL is computed separately, so that the sum of each row is 1. 
For the PL-MoE (Figure~\ref{routing-pl-moe}), most tokens of each PL prefer to be routed to the experts in corresponding group, while a small number of tokens for each PL tend to be routed to the shared expert. For MoE (Figure~\ref{routing-moe}), tokens from all PLs seem to prefer the same set of few experts slightly over the others. PL-MoE has two main advantages over MoE.

First, PL-MoE can not only learn PL-specific knowledge by using the PL-exclusive parameter capacity, but also learn PL-agnostic knowledge with the help of shared parameters. However, the MoE cannot control the allocation of parameter capacity, so the knowledge learned by most experts is mixture of PL-specific and PL-agnostic.

Second, PL-MoE can use expert resources more reasonably to evade the problem of unbalanced expert load. 
% This thanks to the fact that the number of experts in each PL are assigned according to the distribution of PL's resources. However, due to the imbalance of expert load, MoE significantly underperforms PL-MoE in all experiments, and it also does not take full advantage of expert resources. 
More details about efficiency of expert are discussed in the Section~\ref{efficiency}.
% In overall, we observe such a phenomenon: most tokens of each PL prefer to be routed to the experts in corresponding group, while a small number of tokens for each PL trend to be routed to the shared expert. This phenomenon shows that: MultiCoder will use most of data of a PL to learn the knowledge of the PL itself at the PL-MoE layers, while another small amount of tokens are used to learn the generalization knowledge across different PL. This way brings the following two advantages.

% Firstly, MultiCoder not only can learn the transfer knowledge at the bottom layers of model, but also can learn the high-level generalized PL knowledge at the top layers due to the shared expert. Thus MultiCoder can significantly improve the performance of code completion in low-resource PL.

% Secondly, MultiCoder can break through the capacity bottleneck using MoE architecture, and also greatly reduce the interference between different PLs through PL-MoE routing strategy. Therefore, MultiCoder can achieve or even outperform the MonoPL models in high-resource PL.

\subsection{The High Efficiency of PL-level Routing}
\label{efficiency}
\begin{table}[t]\small
    \centering
    \setlength{\tabcolsep}{3pt}
    \renewcommand{\arraystretch}{1.1}
    \caption{The number of experts can be routed in per MoE layer for each PL (including shared expert).}
    \label{tab:experts-PL}
    \begin{tabular}{lccccccc}
    \toprule
    \pmb{Model} & Ruby & Go & JavaScript & PHP & Java & Python & Total \\
    \midrule
    MoE  & 32 & 32 & 32 & 32 & 32 & 32 & 32 \\
    PL-MoE & 3 & 5 & 6 & 7 & 7 & 9 & 32 \\
    \bottomrule
    \end{tabular}
\end{table}

% In this section, we will compare the efficiency of MoE and PL-MoE from the perspective of the number of experts assigned to each PL. 
% The number of experts can be routed in per MoE and PL-MoE layer for each PL as shown in Table~\ref{tab:experts-PL}. 
The number of experts can be routed for each PL as shown in Table~\ref{tab:experts-PL}. 
Overall, the average expert resource occupancy rate of PL-MoE is 19\% of MoE. But PL-MoE outperforms MoE in all experiments. This shows that PL-MoE can efficiently and fully utilize expert resources.

% As mentioned above, the variant of MultiCoder without MoE has two fundamental limitations: the capacity bottleneck and the interference between different PL. While scaling up dense models directly can break through capacity bottleneck, the computational overhead per input also increases proportionally. Thus, this paper proposes to use MoE to scale up MultiPL model without a proportional increase in computational overhead. However, the variant without PL-MoE suffers from the interference between different PLs, due to parameters of experts are shared for all PLs.
% The interference negatively affects both high-resource and low-resource PL. 
% To alleviate this interference as possible, this paper proposes PL-MoE to route different PL separately to different expert groups at the MoE layer. In this way, MultiCoder can effectively alleviate the above two limitations. 

The PL-MoE assigns experts for each PL according to the proportion of data size, while MoE shares all experts to all PLs. 
The most critical limitation of MoE is that it does not distinguish the parameters of each PL clearly enough. 
% In an extreme case, if tokens from different PLs tend to be routed to a small group of experts, there will also be a serious problem of domain knowledge forgetting. 
This not only results in limited improvement of overall performance, but also results in an imbalanced expert load. 
However, PL-MoE allows expert groups to maximize PL-specific knowledge within the group without interference. 
Therefore MultiCoder can further improve overall performance, especially on low-resource languages.

Furthermore, we think more deeply about why the improvement of the variant without PL-MoE is limited on the low-resource PLs. 
We conjecture the reason is that the shared parameters used by the low-resource PLs are not general enough, since most of the experts in the MoE layers are likely to have a mixture of general and PL-specific knowledge. Thus the model may fail to maximize the knowledge transfer toward the low-resource PLs.
% It is hard for ``- PL-MoE'' maximize the transfer knowledge toward low-resource PL.
In contrast, PL-MoE has only a handful of shared experts so that the MultiCoder can selectively route the most general tokens in each PL to the shared experts. And these few shared experts will have strong generalization ability of PLs. 
\section{Related Work}
\label{sec:related}
% \subsection{Code Representation Learning}
% Source code representation learning is a fundamental task for code intelligence. It aims to learn a distributed representation of a source code snippet using deep learning. \cite{iyer2016summarizing} propose Code-NN, which utilizes LSTM~\cite{hochreiter1997long} with an attention mechanism to generate natural language descriptions for C\# and SQL. \cite{alon2019code2vec} propose code2vec, which represents a code snippet as continuous distributed vectors, i.e., code embeddings. To learn the structural information of code, \cite{shido2019automatic} use Tree-LSTM to model structure-style inputs like ASTs. And  \cite{leclair2020improved} use Graph Neural Networks (GNNs) to model these structural features. More recently, many studies focus on utilizing pre-training paradigm to enhance the code representation learning. \cite{feng2020codebert} pre-train CodeBERT by masked language modeling
% and replaced token detection objectives on six programming languages. \cite{ahmad2021unified} pre-train PLBART on an extensive PL-NL corpus to via denoising autoencoding. \cite{guo2022unixcoder} propose UniXcoder, which leverages cross-modal information like ASTs and docstring to enhance code representation.

\subsection{Pre-training for Code Completion}
% For the task of code completion, most studies have leveraged GPT's impressive generation capabilities to improve the quality of code generation. 
\cite{svyatkovskiy2020intellicode} propose GPT-C, which is a variant of the GPT-2~\cite{radford2019language} trained from scratch for multilingual code completion. \cite{lu2021codexglue} pre-train monolingual models named CodeGPT on Python and Java corpus, respectively. 
\cite{liu2020multi} develop a multi-task learning based model named CugLM for code understanding and code generation. 
Most recently, \cite{izadi2022codefill} propose CodeFill that combines learned structure and naming information to achieve multi-token code completion. 
% \cite{kim2021code} pay attention to completing abstract syntax trees (ASTs) by predicting the next node in a flatten tree. 
% \cite{lu2022reacc} propose a retrieval-augmented network for code completion, dubbed ReACC, which exploits external information by retrieving semantically and lexically similar codes from existing codebases.

\subsection{Mixture-of-Experts for efficient pre-training}
% Sparse MoE has been a successful method for scaling pre-training language models to billions of parameters without a proportional increase in training computations~\cite{kudugunta2021beyond}. 
% \cite{shazeer2017outrageously} are the first to prove the effectiveness of MoE in the context of modern deep learning frameworks. 
\cite{lepikhin2020gshard} propose GShard which utilizes the MoE Transformer to dramatically improve machine translation across 100 languages. 
To to mitigrate the problem of complexity, communication costs, and training instability in MoE, \cite{fedus2022switch} propose Switch Transformer, which improves the routing strategy and the experts load balancing loss simultaneously. 
More recently, many researchers focus on how to leverage MoE for efficient multi-domain or multilingual learning. 
% \cite{gururangan2021demix} propose DEMIX layer to explicitly condition on textual domains during language modeling, in order to force experts to specialize to domains. 
And \cite{kudugunta2021beyond} propose task-MoE to improve the Multilingual Neural Machine Translation (MNMT) and actualize inference efficiency.
\section{Conclusion}
\label{sec:conclusion}
In this paper, we propose the MultiCoder to enhance the low-resource code completion via MultiPL pre-training and PL-MoE. 
% In this paper, we delve into the effects of MultiPL pre-training for code completion, in particular on the low-resource programming languages. 
% To this end, the MultiCoder is proposed.
% First, the MultiCoder takes the MultiPL pre-trained GPT as its backbone model.
% Then we introduce the MoE architecture to break through the model capacity bottleneck. Finally, we propose a novel PL-level MoE routing strategy with shared experts to balance the transferability and interference between programming languages. 
The extensive experiments show that the proposed MultiCoder significantly improves the code completion on various programming languages, especially on the low-resource ones. 
We further discuss and analyze the effectiveness of the proposed method. 
% In the future, we will explore the MultiCoder on more downstream tasks. 
% We will also apply PL-MoE to more architectures of pre-trained language models, such as Encoder-only and Encoder-Decoder.

\bibliography{anthology,custom}
\bibliographystyle{acl_natbib}

\newpage
\appendix

% \section{Example Appendix}
\section{Appendix}

\subsection{Model Architecture}
\label{sec:app_model_configs}
\begin{table}[h]\small
    \centering
    \setlength{\tabcolsep}{6pt}
    \renewcommand{\arraystretch}{1.1}
    \caption{Model Architecture.}
    \label{tab:settings}
    \begin{tabular}{lll}
    \toprule
    Parameter & Explanation & Value \\
    \midrule
    $L_{total}$ & Model layers & 12 \\
    $L_{moe}$ & MoE layers & 4 \\
    $h$ & Hidden size & 768 \\
    $S$ & Max sequence length & 1024 \\
    $AH$ & Attention heads & 12 \\
    $E$ & Experts per MoE layer & 32 \\
    $SE$ & Shared experts per MoE layer & 1 \\
    \bottomrule
    \end{tabular}
\end{table}
Table~\ref{tab:settings} provides the architecture details of the pre-trained models in our experiments.

\subsection{Construction Rules of the MultiCC Examples}
\label{sec:app_multicc_rules}
\begin{enumerate}
    \item Tokenize the samples for token-level code completion. 
    \item Add \texttt{<s>} and \texttt{<\textbackslash{s}>} tokens to the beginning and the end of one piece of code respectively.
    \item Replace the line-break \texttt{\textbackslash{n}} as a \texttt{<EOL>} token (end-of-line) for each Python line, since Python does not have ``\}'' or ``;'' to mark the end of a statement like other PLs.
    \item Normalize specific literals as the CodeXGLUE does. The low-frequent literals are replaced by placeholders, e.g. named entities like names, phone numbers and IP addresses. The high-frequent literals are also replaced by placeholders~\footnote{The 200 most frequent strings and 30 most frequent numeric literals are preserved. For example, the common string literal of ``utf-8'' will be normalized by a token of ``<STR\_LIT:utf-8>'', while the uncommon literals are normalized by ``<STR\_LIT>'' or ``<NUM\_LIT>''}. % Sometimes, programmers leave their privacy data in their codes, such as names, phone numbers, IP address. Since we do not encourage models to focus on these string or numeric literals, we normalize these uncommon literals by some special tokens for better user experience. We also consider to common literals, where the 200 most frequent string and 30 most frequent numeric literals are preserved. For example, the common string literal of ``utf-8'' will be normalized by a token of ``<STR\_LIT:utf-8>'', while the uncommon literals are normalized by ``<STR\_LIT>'' or ``<NUM\_LIT>''.
\end{enumerate}

\subsection{Background and Motivation}
\label{sec:background}

\subsubsection{Pre-training for Programming Languages}
Inspired by the remarkable success of pre-training models in the field of NLP, many researchers trend to study the adaptation of pre-trained models to programming languages. One of the earliest attempts is CodeBert~\cite{feng2020codebert}, which is built on BERT~\cite{devlin2019bert} and pre-trained with six common programming languages (including Python, Java, JavaScript, PHP, Ruby, and Go). After that, PLBART~\cite{ahmad2021unified} follows the encoder-decoder architecture and learns multilingual representations of programming and natural language jointly. More recently, GraphCodeBert~\cite{guo2021graphcodebert} introduces two structure-aware pre-training tasks to model the structural information of source code. SYNCOBERT~\cite{wang2021syncobert} utilizes the contrastive learning~\cite{contrastive_learning} with two syntax-guided pre-training tasks to learn better code representation.

% With the rapid development of pre-trained models in programming languages, the progress of code intelligence has been greatly promoted. Some industrial products have already been implemented based on pre-training models, such as Copilot,\footnote{\url{https://copilot.github.com/}} an AI pair programmer which supports features such as code auto completion, converting comments to code, and auto-filling for repetitive code. The launch of these code intelligence products shows that the pre-training paradigm of programming languages is very promising. 

Unfortunately, most of the previous work has focused on mono-programming-lingual pre-training, which performs poorly on the tasks of low-resource programming languages. As mentioned in Section I, insufficient pre-training data for low-resource programming languages limits the performance on these languages. An effective way to improve the performance of low-resource programming languages is multi-lingual pre-training~\cite{devlin2019bert}.

\subsubsection{Multilingual versus Monolingual LMs}
With the success of monolingual English BERT models, the release of multilingual BERT~\cite{devlin2019bert} has led to widespread use of Transformer-based pre-trained multilingual LMs. Guo \textit{ea al.}~\cite{guo2019multilingualparaphrasing} introduces multilingual pre-training into decoder-only models. 
These studies demonstrate that multilingual pre-training models can effectively improve the performance of low-resource languages. 
However, multilingual LMs are still faced with some fundamental limitations~\cite{arivazhagan2019massively}.
% of multilingual LMs are pointed by recent works. 
Conneau \textit{et al.}~\cite{conneau2020unsupervised} find that adding more languages to a fixed model capacity will result in performance drops. This phenomenon is also called the curse of multilinguality (i.e., the lack of ability to represent all languages in a fair way), which can be alleaviated by increasing the model capacity~\cite{chau2020parsing}. Furthermore, Wu \textit{et al.}~\cite{wu2020all} also observe that the monolingual ability of a multilingual model decreases with smaller sizes of pre-training data. These limitations might result in the multilingual models underperforming the monolingual models when applied to a specific language~\cite{virtanen2019multilingual}. 
% For programming language pre-training models, Svyatkovskiy \textit{et al.}~\cite{svyatkovskiy2020intellicode} conduct several experiments on MultiPL models and MonoPL models, and the experimental results show that MultiPL models outperform MonoPL models on low-resource programming languages, while MonoPL models perform better on high-resource programming languages. 

%Although multilingual models can significantly improve the performance on low-resource languages or cross-lingual tasks, their performance loss on other languages is also non-negligible~\cite{johnson2017google,tan2018multilingual}.
Despite recent efforts have been taken to mitigate the limitations, the multilingual LMs are still struggling with how to balance the parameters across different languages~\cite{rust2021good}. To address the problem, we introduce Mixture-of-Experts modules to scale up the model capacity of the MultiPL pre-training.

\subsubsection{Mixture-of-Experts for Scaling Transformers}
The Transformer model~\cite{vaswani2017attention} is widely used in the NLP field,
% natural language processing, 
since it has proven obvious advantages over
% significantly outperforms 
other architectures such as Recurrent Neural Networks (RNNs). A Transformer layer consists of a Multi-Head Self-Attention module and a Feed-forward Network (FFN). Multiple Transformer layers are stacked one by one to constitute a Transformer model. More details about the Transformer model can be referred to~\cite{vaswani2017attention}. The computation of Transformer will increase proportionally with the increase of model parameters. An efficient method to scale up the Transformer is Mixture-of-Experts (MoE)~\cite{lepikhin2020gshard}. A typical MoE framework consists of a gated network and multiple expert sub-networks.
% \yp{, as shown in the upper right of Figure~\ref{fig:PL-MoE}}. 
The gated network calculates the proportion of the output of each expert network for an input $x$, and then adopts a weighted summation method to obtain the final output. The MoE can selectively activate a part of the parameters in the model for different inputs to participate in the calculation, so that the model can be scaled up to billions of parameters without a proportional increase in computation.

\subsubsection{Motivation for the Programming Language Specific Experts (PL-MoE)}
Although the MoE layer expands the model capacity, it does not take into account the exclusive features of different programming languages. In details, all the programming languages share the identical experts on each MoE layer. 
No matter the naive MoE~\cite{lepikhin2020gshard} or the Switch MoE~\cite{fedus2022switch}, the routing strategy of them is trained for token distribution rather than for different programming languages. Therefore, the MoE layers should be improved to guarantee the exclusive features of different programming languages. To protect the exclusive features of different programming languages, we propose a novel PL-level MoE routing strategy (PL-MoE) that provides PL-specific expert parameters. Meanwhile, a set of PL-shared experts are kept so that PL-agnostic knowledge can be learned, e.g. common literals used in various PLs.
% \yp{A visual comparison between the Switch MoE and the proposed PL-MoE is depicted in Figure~\ref{fig:PL-MoE}.}
% The motivation of introducing shared experts is to allow low-resource programming languages to still benefit from transferring knowledge at the MoE layer.

% Lepikhin \textit{et al.} propose Mixture-of-Experts Transformer model, which replaces the dense FFN layer with the MoE FFN layer. More details about MoE transformer are given in~\cite{lepikhin2020gshard}.
% The MoE layer consists of a gating network $\mathcal{G}_{i, E}$ and $E$ FFNs (i.e. expert), such that $\mathrm{FFN}_{1} \ldots \mathrm{FFN}_{E}$. 
% \begin{equation}
%     \mathcal{G}_{i, E}=\operatorname{GATE}\left(x_{i}\right)
%     \label{eqa:1}
% \end{equation}
% \begin{equation}
%     \mathrm{FFN}_{e}\left(x_{i}\right)=w o_{e} \cdot \operatorname{ReLU}\left(w i_{e} \cdot x_{i}\right)
%     \label{eqa:2}
% \end{equation}
% \begin{equation}
%     y_{i}=\sum_{e=1}^{E} \mathcal{G}_{i, e} \cdot \mathrm{FFN}_{e}\left(x_{i}\right)
%     \label{eqa:3}
% \end{equation}
% where $x_{i}$ is the input token at position $i$ to the MoE layer. The vector $\mathcal{G}_{i, E}$ is computed by the gating network. And it has one non-negative value for each expert, which determines the assignment of experts. And the ${FFN}_{e}$ (an expert) refers to a 2-layer fully-connected network using ReLU activation function, $w i_{e}$ and $w o_{e}$ refer to the input and output projection matrices for the $e$-th expert. 

\subsection{The Performance of MultiCoder during Pre-training Phase}
\label{pre-training}
\begin{figure}[t]
    \centering
    \includegraphics[width=6.5cm]{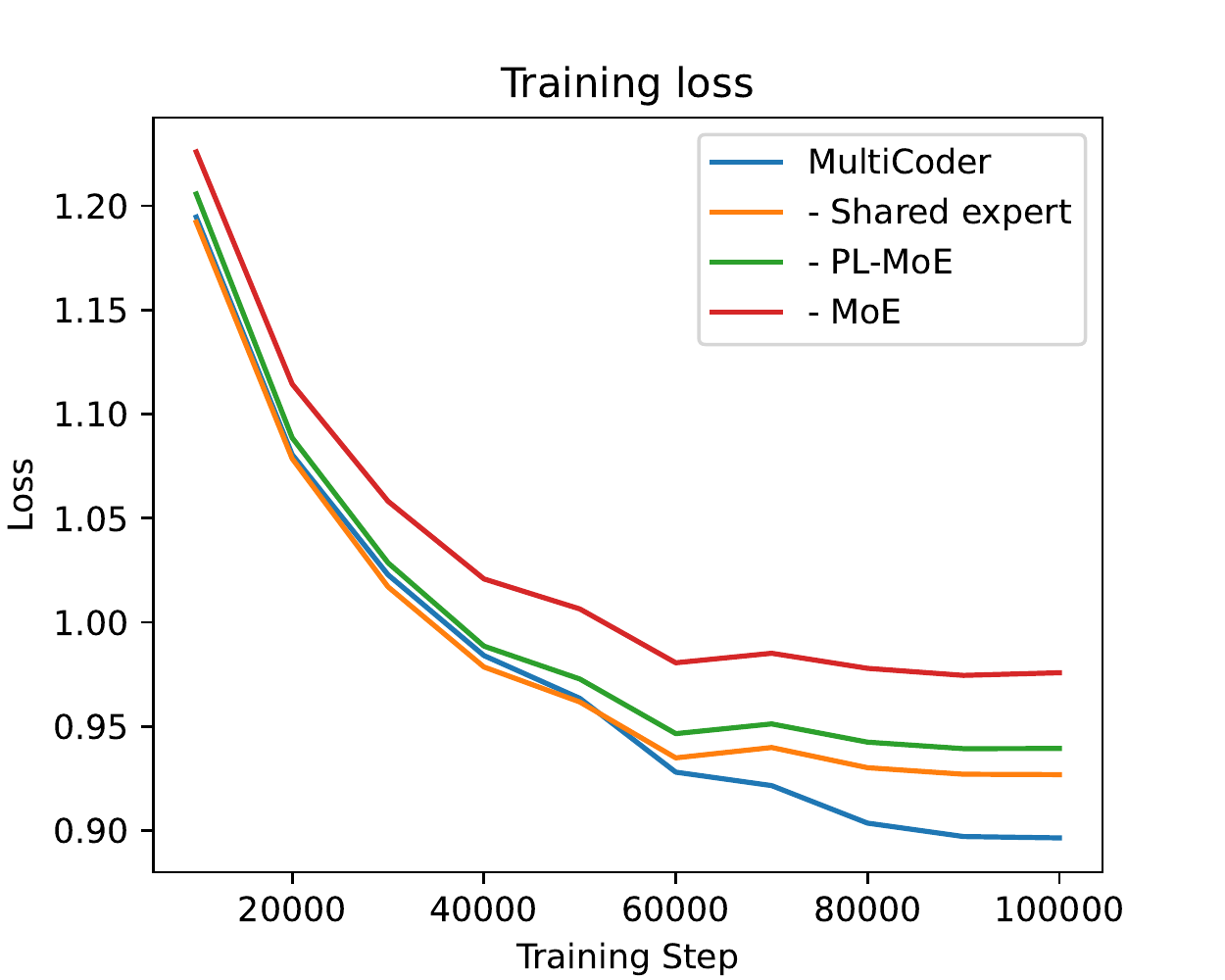}
    \caption{Training loss of language models.}
    \label{fig:train-loss}
\end{figure}
\begin{figure}[t]
    \centering
    \includegraphics[width=7.5cm]{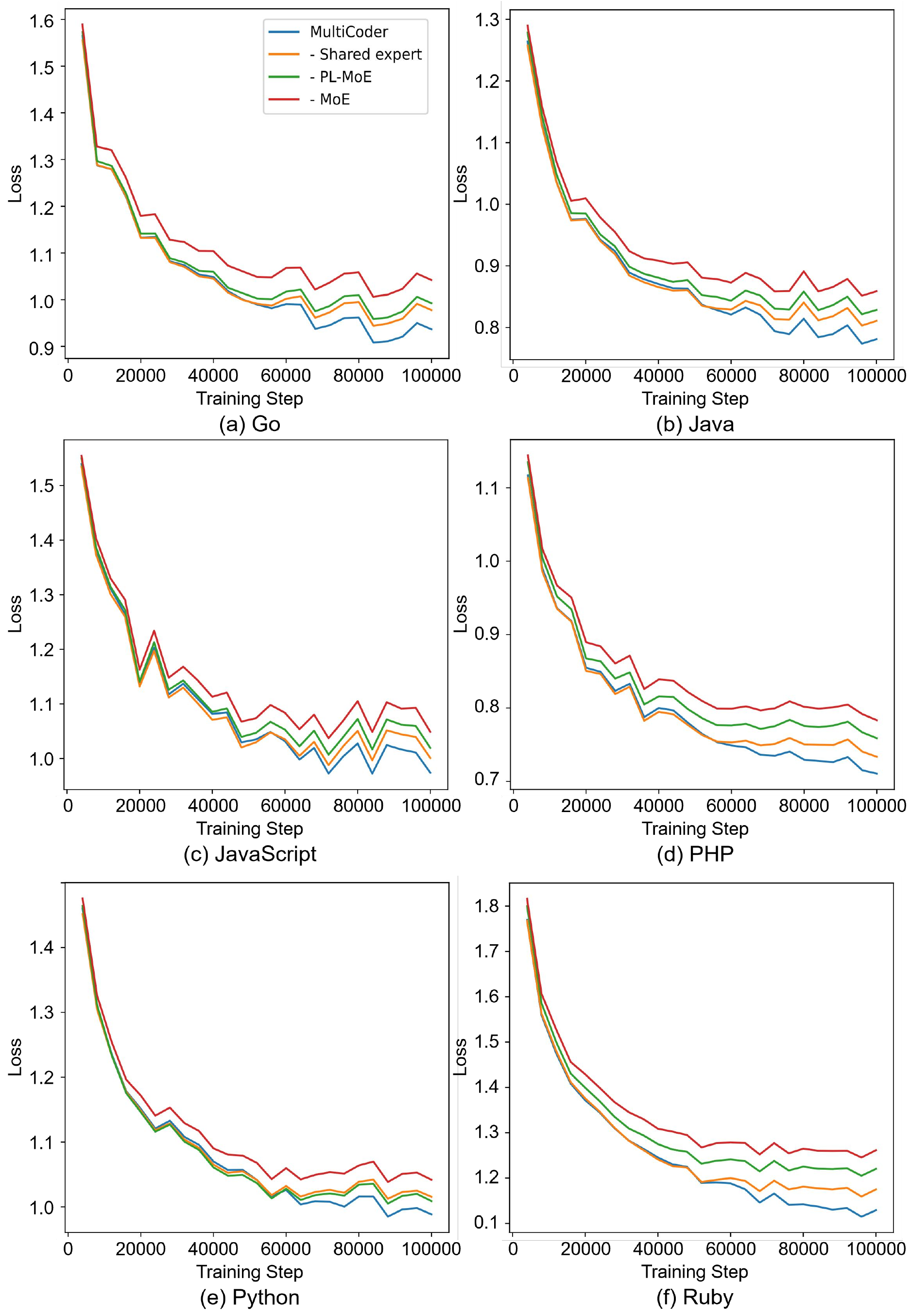}
    \caption{Validation loss of language models on different PLs.}
    \label{fig:valid-loss}
\end{figure}

In this section, we focus on the performance of the MultiCoder during pre-training phase. To facilitate analysis, we plot the loss function curves of the models on the training and validation sets during pre-training in Figure~\ref{fig:train-loss} and \ref{fig:valid-loss}, respectively, where the curves on the validation set are plotted separately for different PLs. 

% As shown in Figure~\ref{fig:train-loss}, we observe that the values of loss curve of ``MultiCoder'' are the lowest, followed by ``- Shared expert'', and the highest is ``- MoE''.
As shown in Figure~\ref{fig:train-loss}, MultiCoder outperforms its variants without MoE by nearly 10\% in the pre-training stage, and outperforms other variants by at least 4\%.
The results show that MultiCoder converges the best on the pre-trained dataset, and proves again the effectiveness of the PL-MoE and the shared expert. Furthermore, all the models with MoE outperform the variant without MoE. 
% This phenomenon demonstrates that the benefits of using MoE to break through the capacity bottleneck can also be realized during the pre-training phase. 
This illustrates that using MoE to break through the capacity bottleneck can also bring benefits in the pre-training stage.
% In addition, the figure also demonstrates the advantages of introducing shared experts in the PL-MoE layer. 

On the other hand, as shown in Figure~\ref{fig:valid-loss}, the MultiCoder consistently has the lowest validation loss on all the PLs. It depicts that the MultiCoder maintains a strong generalization ability on the validation set while fitting the training set well. 
% Especially in low-resource PL, such as Ruby, MultiCoder outperforms ``- MoE'' by nearly 13\%, there are the most significant gaps in loss values between ``MultiCoder'' and other models.
Especially in the low-resource PLs, there are more significant gaps in loss values between the MultiCoder and its variants. For Ruby, the MultiCoder outperforms the variant without MoE by nearly 13\%.
We further find a trend that the smaller the pre-training set is, the greater advantage the MultiCoder obtains. 

\subsection{Threats to Validity}
\label{threats}
We have identified the main threats to the validity of our approach as follows:
\begin{itemize}
    \item \textbf{The number of downstream tasks.} Due to the restriction of GPT-style backbone model, we only select the code completion as the downstream task. Although code completion is a representative generation task of code, others tasks also can verify the effectiveness of the model in the code generation ability, such as text-to-code generation. In our future work, we will evaluate the effectiveness of MultiCoder with more generation tasks of code.
    \item \textbf{The selection of evaluation metrics.} In this paper, we choose the token-level accuracy and the Levenshtein distance as our evaluation metrics. However, they both focus on fine-grained assess, i.e., token-level and character-level. There are some evaluation metrics focus on more coarse-grained, such as exact match accuracy (EM). We will use more evaluation metrics of different granularity to verify the code generation quality of MultiCoder.
    \item \textbf{The selection of backbone model.} We choose GPT-2 as the backbone of MultiCoder in this paper. Actually, the PL-MoE routing strategy is model-agnostic so that it can be applied on others backbone, such as Encoder-only and Encoder-Decoder models. In the future, we will try to apply the PL-MoE to BERT and T5, and verify the robustness of PL-MoE in different architectures.
\end{itemize}
% \label{sec:appendix}

% This is a section in the appendix.

\end{document}